\documentclass[]{ceurart}

\usepackage{longtable}
\usepackage{enumitem}
\usepackage{hyperref}
\usepackage{csquotes}
\usepackage{changepage}
\usepackage{listings}
\usepackage{multicol}
\usepackage{pifont}
\usepackage[labelfont=bf]{caption}
\usepackage{multicol}
\usepackage{makecell}

\newcommand{\listingcaption}[1]%
{%
\refstepcounter{lstlisting}\hfill%
Code \thelstlisting -- #1\hfill
}%

\begin{document}

\copyrightyear{2023}
\copyrightclause{Copyright for this paper by its authors.
 Use permitted under Creative Commons License Attribution 4.0
 International (CC BY 4.0).}

\conference{2nd international workshop
KM4LAW – Knowledge Management and Process Mining for Law, July 17-20, 2023 (Sherbrooke, Québec, Canada)}

\title{An automated method for the ontological representation of security directives}

\author[1]{Giampaolo Bella}[
orcid=0000-0002-7615-8643,
email=giamp@dmi.unict.it,
]
\address[1]{Università degli Studi di Catania}

\author[1]{Gianpietro Castiglione}[%
orcid=0000-0003-2215-0416,
email=gianpietro.castiglione@phd.unict.it,
]

\author[1]{Daniele Francesco Santamaria}[%
orcid=0000-0002-4273-6521,
email=daniele.santamaria@unict.it,
]

\begin{abstract}
Large documents written in juridical language are difficult to interpret, with long sentences leading to intricate and intertwined relations between the nouns.
The present paper frames this problem in the context of recent European security directives.
The complexity of their language is here thwarted by automating the extraction of the relevant information, namely of the parts of speech from each clause, through a specific tailoring of Natural Language Processing (NLP) techniques. These contribute, in combination with ontology development principles, to the design of our automated method for the representation of security directives as ontologies. 
The method is showcased on a practical problem, namely to derive an ontology representing the NIS 2 directive, which is the peak of cybersecurity prescripts at the European level.
Although the NLP techniques adopted showed some limitations and had to be complemented by manual analysis, the overall results provide valid support for directive compliance in general and for ontology development in particular.
\end{abstract}


\maketitle
\section{Introduction} \label{sec:intro}
The promulgation of security directives is a new and increasingly used approach to addressing security issues on a large scale. As the capillarity of security risks rises, so does the complexity of the security directives. In consequence, both the interpretation and the compliance check of each of the defined measures are highly prone to inconsistencies. This forms the motivation for the work presented in this paper, which illustrates a method to treat, and ultimately apply security directives by making them more widely accessible with respect to the legal context from which they originate.
Our method innovatively combines techniques and tools from two known branches of Informatics: natural language processing (NLP) and ontology development. 

To introduce the basics of both constituent branches, we assert the main definitions.
NLP is a technique for making the human natural language more and more understandable for machines, hence for automatic processing. NLP is widely used in the general context of Artificial Intelligence, and it can leverage machine learning algorithms to attain better performance.
Then, \emph{``An ontology is an explicit specification of a conceptualisation''} \cite{GRUBER1995907}, hence an ontology allows us to model a domain through the definitions of its entities and axioms involving them. In particular, entities include classes (categories), individuals (instances of classes) and properties (relations among individuals). Ontologies inherit the logical reasoning capabilities from description logic (DL), a family of formal and decidable formalism tailored towards representing terminological knowledge of a domain in a structured and well-understood way. An advantage of ontologies is that they offer mathematically sound support that can be  useful for the compliance verification phase.
We apply our method to the NIS 2 Directive (Directive (EU) 2022/2555) \cite{nisPDF}, which was promulgated by European Commission on 16 January 2023 and can be considered the main European framework for cybersecurity. It aims to achieve a high common level of security across the EU.

Our method promotes ontology engineering and development towards the precise \emph{representation} of the Directive in support of automated compliance checking. Meanwhile, it employs specific NLP techniques for the grammatical tagging of the Parts Of Speech (POS) to automate the extraction of the relevant information from the Directive to be represented within the ontology. 
More precisely, by following the established Methontology methodology \cite{Methontology}, the ontology is developed as a Python program leveraging the RDFLib library~\cite{rdflib} for what concerns the manipulation of ontological artefacts; the NLP part is developed by using ClausIE~\cite{claucy}, a sub-library of Spacy~\cite{spacy}. Our NIS 2 ontology is the main result of this paper, besides the general method used to produce it. Notably, the NLP techniques that we used worked generally well but were put at stake by the lengthy style of the Directive. Therefore, we also conducted a manual clause analysis next to the NLP one, indicating the limitations of the latter. The experiments shown in this paper are online available \cite{gitNIS}.

An overview of related work is illustrated below (Section \ref{sec:rw}), while the description of our method is illustrated in Section \ref{sec:onto-nlp} and its application to the NIS 2 as a relevant case study in Section \ref{sec:demo-method}. The evaluation of the method is illustrated in Section \ref{sec:eval} and the Conclusions are finally drawn (Section \ref{sec:conclusions}).

\section{Related work} \label{sec:rw}
For the representation of legal documents through ontologies, the works that have been produced are not many. 
In a similar context to the NIS 2 Directive, some work has been done on GDPR; in particular, Sawasaki T. et al.~\cite{DBLP:conf/semweb/SawasakiST22} leverage not ontological reasoning for resolving legal disputes on personal data transferring between different jurisdictions, in this case between Europe and Japan, while Pandit et al. \cite{Pandit2018GDPRtEXTG} propose GDPR as a linked data resource, leveraging pre-existing vocabularies for concepts linking.

A remarkable software product developed for leveraging NLP for compliance verification is the CyberStrong platform \cite{cyberstrong}. 
However, it does not adopt ontological reasoning, and it is difficult to view how it works since it is a commercial product.
The integration of ontologies and NLP has been considered in a few works. In some contexts, ontologies have been used for supporting NLP in the phases of information retrieval \cite{SeseiOnto} \cite{articleEstival}.

The closest work to ours was published by Zhang J. et al. \cite{doi:10.1061/(ASCE)CP.1943-5487.0000346}. The use case proposed is emblematic since the integration of ontologies and NLP is used for compliance verification purposes. However, the information extraction is not applied in a security context, and also in this case, the ontologies are considered as a support for NLP and not for representation of juridical language, as in our case.

TextRazor is a commercial tool with an online interface and also API access offering valid support for POS tagging~\cite{textR}. It is clear that its structured, intensive use requires the purchase of an API key, which future researchers would have to fetch for their own usage. Because we aim at providing a freely accessible tool to automate the tasks of anyone who pursues the ontological representation of security directives, it means that we must appeal to existing open-source products and develop them as necessary.

Among the open source products that can be leveraged for POS tagging comes CoreNLP, a toolkit developed at Stanford \cite{manning-EtAl:2014:P14-5} that appears to be valid for POS tagging, hence worth evaluating for our purposes in the future. While CoreNLP requires the setup of a server to route requests to the Spacy library can be easily leveraged by a calling program, and works well to name the leading entities in a statement, its clauses and its parts of speech~\cite{spacy}.
We see below that this library is chosen as a basis for our developments.

\section{Automated method to represent directives as ontologies}\label{sec:onto-nlp}
This Section is the core of the present work and describes our method to source information from a given security directive and leverage it, through precise modelling decisions, to build an ontology in an automated fashion. Notably, the method takes advantage of the use of specific NLP techniques for POS tagging and then sets out to adapt the various steps for building an ontology on our particular application, as well as to develop the corresponding software. Therefore, we initially have to establish what methodology of ontological representation to adopt and use as a consolidated basis for our developments.

At an abstract level, we take a \emph{waterfall} approach, which prescribes a sequence of steps such as specification, through implementation and up to evaluation, as with any software.
A more specific incarnation of the waterfall approach is the Methontology methodology~\cite{Methontology} of ontological development, which consists of precise steps: 1) \emph{Specification}, 2) \emph{Knowledge Acquisition}, 3) \emph{Conceptualisation}, 4) \emph{Integration}, 5) \emph{Implementation}, 6) \emph{Evaluation}, and 7) \emph{Documentation}. Our method leverages Methontology and, 
in particular, step 3 can be expected to be crucial for our purposes, for example, to fit the NLP part with the rest of the developments and produce an ontology in the end. By contrast, we shall see that the other steps need to be tailored towards the representation of a security directive. 
The following subsections illustrate the various steps of our method.

\subsection{Designing a modular ontological pattern for security directives} 
This step is inspired by step 1 of Methontology.
Because cybersecurity is increasingly challenging worldwide, more and more legislation about it is appearing. Therefore, virtually every organisation is facing the problem of interpreting that legislation before a compliance process can be initiated. 

\begin{figure}[ht]
  \centering
  \includegraphics[scale=0.7]{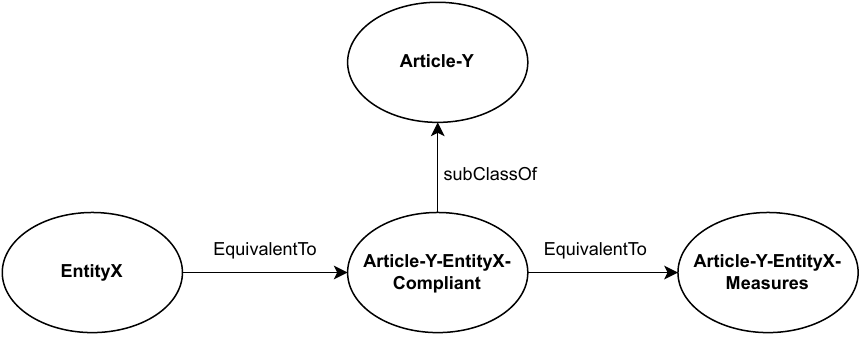}
  \caption{Overview of the main classes for the security directives}
  \label{fig:ontological-template}
\end{figure}

A specific category of European legislation is represented by security directives.
In particular, we want to design a modular structure of the security directives, in which, assuming they are divided into articles, every \emph{Article} can be represented as a class  as depicted in Fig. \ref{fig:ontological-template}, where a node of the diagram represents a class and an edge stands for a class axiom. The core idea is to leverage ontological reasoning to check that an entity follows the measures created specifically for it, and for that, the exact relations between modules must be established. 

The compliance check will be carried out by one of the possible entities named in the security directives, here conceptualised as \emph{EntityX}. In fact, the term entity has two meanings: an ontological entity and the entity defined within the context of a security directive. Every EntityX, for example, a NIS 2 Member State or a GDPR Data Protection Officer (DPO), will be associated with a specific sub-article class, namely \emph{Article-Y-EntityX-Compliant}, which describes the specific measures Article-Y refers to EntityX. For example, Article 10 of NIS 2 Directive involves entities Member State and CSIRT (Computer Security Incident Response Team), and consequently, we will have entity Article-10 with subclasses Article-10-MemberState-Compliant and Article-10-CSIRT-Compliant classes, each in turn with their respective related classes.

\begin{figure}[ht]
  \centering
  \includegraphics[scale=0.7]{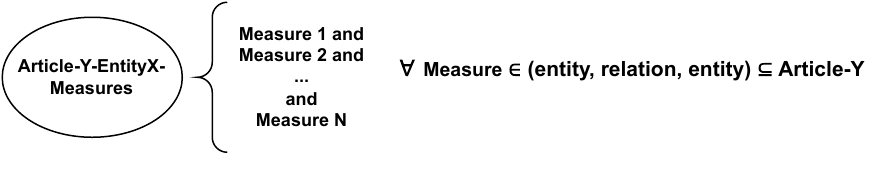}
  \caption{Representation of articles}
  \label{fig:articles-template}
\end{figure}

The specification of measures is represented by module \emph{Article-Y-EntityX-Measures}, connected to Article-Y-EntityX-Compliant with the EquivalentTo property so that there is direct equivalence between an article and the measures defined therein. It represents the measures through the combination of ontological entities, individuals, classes and properties; in particular, we want to design the measures of each Article as a logical conjunction of exactly each action to be done by EntityX. Fig. \ref{fig:articles-template} exemplifies this concept.

If the relation is  \emph{EquivalentTo},  by the transitive property we obtain that EntityX is equivalent to Article-Y-EntityX-Measures. 
The meaning of this association brings its benefits in a possible reasoning phase. For example, let us model an individual (namely a specific instance of a class) \emph{EntityX-test} of class Entity-X for the sake of verifying compliance with Article-Y. Then, only if EntityX-test owns all the properties related to the specific Article-Y, can the reasoning  infer that EntityX-test equals EntityX. This in turn means that EntityX-test is compliant with Article-Y. The term compliant is specifically chosen for this phase, in fact, if the individual possesses all the measures, then the inference  ``EntityX-test inferred Article-Y-EntityX-Compliant'' will be produced.

By adopting this type of design, we can represent the security directives with small and separated modules, and this facilitates the management of the ontology and of the directives themselves. The separation between entities and their respective measures aims at a loose coupling in support of readiness for modifications if any errors are found or modifications become necessary.
The details of how each measure will be represented will be discussed later.

\subsection{Extraction and tagging of parts of speech from security measures}
Once the modular ontological pattern is available, we continue to the knowledge acquisition step, which is inspired by step 2 of Methontology but is here tailored to the parts of speech.

POS tagging involves, in particular, the extraction of essential parts of speech such as subject, predicate and object. By framing these in the context of security directives, we have that the subject is the NIS 2 entity to which a measure is referred, the predicate is the action the entity must perform, and the object is the entity or composition of nouns upon which the action is directed. 

There are two main approaches to the  automatic extraction of parts of speech, namely to entirely rely on some full-fledged software as an oracle or to resort to dedicated coding where fine-tuning is arguably possible. As discussed above (\S\ref{sec:rw}), we pursue the open source philosophy, hence we take the second approach and, in particular, leverage Spacy to develop a parser that takes the given security directive as input and outputs the POS tagging based on the sub-class ClaucIE \cite{claucy} of Spacy. The present step prescribes, in turn, the following ones.

\paragraph{Preprocessing}
Juridical language may take various complicated structures. A recurrent structure sees a sentence with various, alternative objects (or objective clauses). The various objects are itemised, while the subject and verb are omitted to limit redundancy. One of the preprocessing routines is to introduce that redundancy, namely to build the full sentence for each object in order to facilitate the next steps. Other preprocessing routines are minor and omitted here.

\paragraph{Identification of sentences}
We analyse each sentence that grammatically ends with a full stop, also taking advantage of the preprocessing just discussed. This choice fits perfectly with our goal of extracting the parts of speech, as we shall see in the next sub-step. 

\paragraph{Grammatical tagging of each sentence}
Each extracted sentence is subsequently subjected to functions for the tagging of the relevant parts of speech. 
One of the biggest complicating factors can be anticipated to be the high complexity of some sentences of the security 
directive given by input. Sometimes, sentences may span over many lines and may adopt a somewhat intricate language. 
Therefore, we appeal to the specific features of  the Spacy library that turn out useful at this particular point, and we leverage ClausIE, the sub-library of Spacy oriented at the identification of clauses. 
The advantage of using ClausIE instead of Spacy is that dividing the single sentence into several clauses can be expected to raise the probability of obtaining the correct and wanted results in terms of tagging.

By applying the functions of ClausIE, each sentence is parsed and the identified pattern is generated. The pattern is a combination of the main POS, namely Subject (S), Verb (V), Object (O), Complement (C) and Adverb (A).
Two more considerations need to be made here. One is that POS Verb refers to what we introduced as the predicate, the other one is that we consider POS Object and Complement with the same role. 
However, our experiments demonstrate that not all clauses can be automatically extracted, particularly when a number of clauses are listed, each depending on a previous one. 
In such cases, which we shall evaluate below on practical demonstration of our method, the object complement of a sentence is the entire part of the sentence following the main verb. 

\paragraph{Tabulation}
The patterns of tags produced provide us with the information we need for the next steps. For this reason, we now take care of storing them in tables for possible statistical analyses to be conducted later or simply to support the subsequent development of the target ontology.

However, it cannot be assumed that our tool succeeds in finding all parts of speech in all cases, namely it might be the case that its output is either incomplete or entirely wrong. In such cases, we as analysts shall be prepared to step in and perform manual tagging. In particular, we also did manual tagging of all relevant articles of the NIS 2, as we shall see in the next Section while our method is demonstrated.

\subsection{Mapping of grammatical tags onto modular ontological patterns}
The information collected with tables in the previous step of our method supports the present step. Inspired by step 3 of Methontology, here we instantiate the patterns defined in the first step by means of the extracted parts of speech. This conceptualisation of the domain is fundamental to consistently executing the next phase, where the target ontology is implemented. 
Therefore, this step offers the first conceptual representation of the input directive and assumes a concrete meaning in the application of the method.

\subsection{Implementation of an automated ontology builder}
This implementation step, inspired by step 5 of Methontology, consists of automating the ontology-building process for the specific parts of the security directive identified by means of POS tagging.
The implementation of the ontology is divided into a manual phase and an automatic phase. The first consists of manually defining  with Protégé \cite{DBLP:journals/aimatters/Musen15} some entities that cannot be automatically extracted,  but this lies outside the focus of the present paper. 

The automatic phase leverages the code we developed for the automated representation with the support of the RDFLibrary \cite{rdflib}. Our code contains three main modules: module \textit{i} to parse the document taking the exact articles and sentences, module \textit{ii} to apply NLP calls to each extracted part and store the results, and module \textit{iii} to take the stored results and give them an ontological representation. 
\lstset{
   language=Python,
   breaklines=true,
   numbers=left,
   showstringspaces=false,   
   xleftmargin=\parindent,
    basicstyle={\small},
   prebreak = \raisebox{0ex}[0ex][0ex]{\ensuremath{\hookleftarrow}}
}
\begin{lstlisting}[stepnumber=1,multicols=2,  label={code-onto}]
def create_ontology(subj, pred1, obj1, pred2, obj2):

g = Graph()
ex = Namespace("http://nas.onto/")

a = BNode()
b = BNode()
c = BNode()

mainClass = URIRef("http://nas.onto/Article-X-Compliant")

subject = URIRef("http://nas.onto/" + subj)
relation1 = URIRef("http://nas.onto/" + pred1) 
object1 = URIRef("http://nas.onto/" + obj1)  
relation2 = URIRef("http://nas.onto/" + pred2) 
object2 = URIRef("http://nas.onto/" + obj2)   

g.add((a, RDF.type, OWL.Restriction))
g.add((a, OWL.onProperty, relation1))
g.add((a, OWL.someValuesFrom, object1))

g.add((b, RDF.type, OWL.Restriction))
g.add((b, OWL.onProperty, ex.relation2))
g.add((b, OWL.someValuesFrom, ex.object2))

coll = BNode()
Collection(g, coll, [a, b])

g.add((c, OWL.intersectionOf, coll))
g.add((c, RDF.type, OWL.Class))

g.add((mainClass, OWL.equivalentClass, c))
g.add((mainClass, RDF.type, OWL.Class))

g.add((subject, OWL.equivalentClass, mainClass))
\end{lstlisting}
\listingcaption{Extract of python code for the automated creation of the measures of the Directive on ontology\label{code-onto}}
\vspace*{0.5cm}

For example, an extract from module \textit{iii} is shown in Code \ref{code-onto}. 
The code exemplifies the general case of an Article, treating only two measures of it. 
No imports of the library are shown to lighten the illustration. The imports used are submodules of the RDFLib library.
The main function defined on line 1 has input subj as EntityX of the modular pattern, and then has pred1, obj1, pred2, obj2 as the extracted POS for the representation of the measures.
Each of them is associated with a specific ontological entity through a reference via URI (lines from 12 to 16). However, the graph representing the ontology and the namespace must be created first (lines 3 and 4).

As we want each measure to be considered essential for compliance checking, we defined them as ontological restrictions (lines 18 to 20, and from 22 to 24), which are assigned to the ontological nodes a and b, defined in lines 6 and 7.
Each restriction involves a specific predicate (or verb, here representing a relation) and object. The relation is represented by an ontological property. The creation of the restriction will output the result \emph{<<predicate>> some <<object>>} where \textit{some} is an ontological keyword necessary for defining an existential restriction, namely classes of individuals who participate in at least one relationship along a specified property.
Once the restrictions have been defined, it is necessary to connect them together in order to obtain their logical conjunction.  
An exemplification of this would result in \emph{(<<predicate1>> some <<object1>>) and (<<predicate2>> some <<object2>>) and ... and (<<predicateN>> some <<objectN>>)}.

The logical conjunction implements a correct compliance check, namely only if all the restrictions are simultaneously respected, then there will be compliance with the relevant article.
We connect the restrictions with the constructor Collection and we associate the collection to the node c (defined in line 8) through \emph{OWL.intersectionOf} (lines from 26 to 30). The last step consists of associating the now ontological measures within the subject. In lines 32 and 33, we associate the measures (contained in c) to the correct article (mainClass, which is defined in line 10), and second, we create a specific class for it. 
The Article (mainClass) is, in the end, associated with the subject, in order to link the measures that it contains to the specific entity to which they are addressed (line 35). 

\section{Demonstration of our method on the NIS 2 directive} \label{sec:demo-method}
This Section demonstrates our automated method to represent directives as ontologies in the particular case of the latest European directive on cybersecurity, the NIS.

\subsection{Designing a modular ontological pattern for security directives}
\begin{figure}[ht]
  \centering
  \includegraphics[scale=0.7]{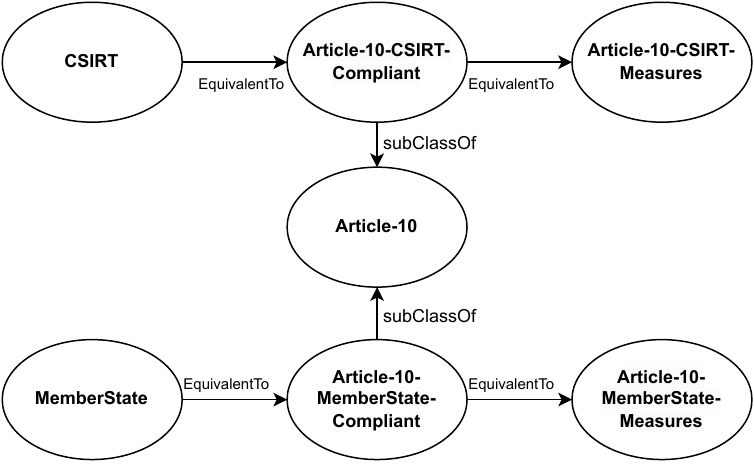}
  \caption{Pattern applied to an extract of Article 10}
  \label{fig:nis2-pattern-art10-bis}
\end{figure}

Following the general guidelines given above through the description of our method, this step prescribes the definition of the modular ontological pattern for the NIS 2. For example, Fig.~\ref{fig:nis2-pattern-art10-bis}, presents the pattern for an extract of Article 10.

\subsection{Extraction of fundamental entities and relations from security measures}
The parts of the Directive subject to acquisition are only those in which the specific measures are defined for regulatory bodies to follow and, in particular, articles from 7 to 37. The remaining articles are not relevant to our purposes because they pertain to premises, definitions and abbreviations, hence do not describe specific compliance actions or activities for the various entities to conduct.

Let us take here the following extracts, respectively from Article 8 and 14 of the Directive, as examples: 
\begin{displayquote}
5. Member States shall ensure that their competent authorities and single points of contact have adequate resources to carry out, in an effective and efficient manner, the tasks assigned to them and thereby fulfil the objectives of this Directive. 

3. The Cooperation Group shall be composed of representatives of Member States, the Commission and ENISA. The European External Action Service shall participate in the activities of the Cooperation Group as an observer. The European Supervisory Authorities 
(ESAs) and the competent authorities under Regulation (EU) 2022/2554 may participate in the activities of the Cooperation Group in accordance with Article 47(1) of that Regulation 
\end{displayquote}

Preprocessing is empty in this case. For what concerns the identification of sentences, we get respectively 1 and 3 sentences from the two extracts.
Then, we run our parser and obtain the following structured extracts:

\begin{lstlisting}[caption=Extract of application of ClausIE to Article 8, label=lst:out1]
[<SVOC, Member States, shall ensure, None, the tasks assigned to them and thereby fulfil the objectives of this Directive, that their competent authorities and single points of contact have adequate resources to carry out, in an effective and efficient manner,, []>, <SVO, their competent authorities and single points of contact, have, None, adequate resources to carry out, None, [in an effective and efficient manner]>]
\end{lstlisting}

\begin{lstlisting}[caption=Extract of application of ClausIE to Article 14, label=lst:out2]
[<SV, The European External Action Service, shall participate, None, None, None, [in the activities of the Cooperation Group, as an observer]>] Clauses [`The European External Action Service shall participate in the activities of  the Cooperation Group as an observer', `The European External Action Service shall participate in the activities of  the Cooperation Group', `The European External Action Service shall participate as an observer']
\end{lstlisting}

The optimal case would be to obtain a pattern that has extracted all relevant POS, namely at least SVOC. However, the complexity of sentences can be problematic in general for the Spacy library, and our parser may either output a partial set of POS or suggest more than one possible tagging for the same clause. In such cases, we obviously have to intervene and finalise the tagging manually
For example, analysing code extracts \ref{lst:out1} and \ref{lst:out2}, it is clear that the first pattern is SVOC, is correct and can be considered complete, while the second is correct but not complete because our tool was unable to identify the object. In this case, the POS will still be collected, as the ontology must cover the entirety of articles 7-37 of the Directive, but we must review all POS and extract the missing ones manually.


Iterating the previous step to all the selected parts of the Directive, a classification of all ontological-related information is then collected in tables. The information we collected suggests how much we are able to fully automate the creation of an ontology. 
For each article, we build a table to represent the information that was extracted.
For each sentence number in the article, we take note of subject, verb and object. For each of these three POS, we have two versions, namely the correct one that we identified manually and the version derived through our parser, which we prefix with an ``I''. 
For example, the subject that was identified manually is headed as ``Sub'' and the version that was identified automatically is headed as ``I-Sub''. Moreover, there is a third column, identified by the suffix ``HIT'' for each tag to identify whether the tool worked well or not. 
All symbols are explained Table \ref{tab:legend}. More precisely, the tick can identify a correct, wrong or incomplete extraction --- we consider the incomplete extraction as correct.
Conversely, a cross symbol means that we found the automatic tagging to be wrong.
The same considerations are valid for verb and object. Additionally, the objects are collected in the tables reachable via the references in Appendix.

\begin{table}[h]
  \begin{minipage}{.5\linewidth}
    \centering
    \begin{tabular}{|c|c|}
      \hline
      \textbf{Name} & \textbf{Acronym}\\
      \hline
      Member State & MS  \\
      \hline
      National Cybsersecurity Strategy & NCS \\ 
      \hline
      Competent Authority & CA \\ 
      \hline
      Commission & Co \\
      \hline
      Point of Contact & POC \\
      \hline
      CSIRT & C \\
      \hline
      ENISA & E \\
      \hline
      Cooperation Group & CG \\
      \hline
      The European External Action Service & EEAS \\
      \hline
      European Supervisory Authorities & ESA \\
      \hline 
      CN & CSIRTs Network \\
      \hline
      Eu-Cyclone & EuC \\
      \hline
    \end{tabular}\par
  \end{minipage}%
  \begin{minipage}{.7\linewidth}
    \centering
    \begin{tabular}{|c|c|}
      \hline
      \textbf{Symbol} & \textbf{Acronym}\\
      \hline
      \ding{52} & Correct answer \\
      \hline 
      \ding{56} & Wrong answer \\
      \hline 
      \ding{121} & Correct but incomplete answer \\
      \hline 
      P & Passive Verb \\
      \hline
    \end{tabular}\par
  \end{minipage}
  \caption{Legend}
  \label{tab:legend}
\end{table}

\begin{table}[h]
\begin{adjustwidth}{0cm}{}
  \begin{tabular}{|l||c|c|c||c|c|c||c|c|c|}
  \hline
  N. & Sub & I-Sub & Sub-HIT & Verb & I-Verb & Verb-HIT & Obj & I-Obj & Obj-HIT \\
  \hline 
  1 & MS & MS & \ding{52} & designate & designate & \ding{52} & \ref{n8-1-1} & \ref{s8-1-1} & \ding{121} \\
  \hline
  2 & CA & CA & \ding{52} & monitor & referred & \ding{56} & \ref{n8-2-1} & \ref{s8-2-1} & \ding{121} \\
  \hline
  3.1 & MS & MS & \ding{52} & designate & designate & \ding{52} & \ref{n8-3-1} & \ref{s8-3-1} & \ding{121} \\
  3.2 & CA & CA & \ding{52} & be & be & \ding{52} & \ref{n8-3-2} & \ref{s8-3-2} & \ding{56} \\
  \hline
  \dots & \dots & \dots & \dots & \dots & \dots & \dots & \dots & \dots & \dots \\
  \hline
  5 & MS & MS & \ding{52} & ensure & ensure & \ding{52} &  \ref{n8-5-1} & \ref{s8-5-1} & \ding{52} \\
  \hline
  6.1 & MS & MS & \ding{52} & notify & notify & \ding{52} &  \ref{n8-6-1} & \ref{s8-6-1} & \ding{121} \\
  \hline
  \dots & \dots & \dots & \dots & \dots & \dots & \dots & \dots & \dots & \dots \\
  \hline
  \end{tabular}
  \caption{Article 8 POS table}
  \label{tab:tab-art-8}
\end{adjustwidth}
\end{table}

\begin{table}[h]
\begin{adjustwidth}{-1.5cm}{}
\begin{tabular}{|l||c|c|c||c|c|c||c|c|c|}
  \hline
  N. & Sub & I-Sub & Sub-HIT & Verb & I-Verb & Verb-HIT & Obj & I-Obj & Obj-HIT \\
  \hline 
  1 & - & - & - & P & P & - & \ref{n14-1-1} & \ref{s14-1-1} & \ding{56} \\
  \hline
  2 & CG & CG & \ding{52} & carry & carry & \ding{52} & \ref{n14-2-1} & \ref{s14-2-1} & \ding{52} \\
  \hline
  3.1 & CG & CG & \ding{52} & P - composed & P - composed & \ding{52} & \ref{n14-3-1} & \ref{s14-3-1} & \ding{56}\\
  3.2 & EEAS & EEAS & \ding{52} & participate & participate & \ding{52} & \ref{n14-3-2} & \ref{s14-3-2} & \ding{56}\\
  3.3 & ESA - CA & ESA & \ding{121} & participate & participate & \ding{52} & \ref{n14-3-3} & \ref{s14-3-3} & \ding{56} \\
  \hline
  \dots & \dots & \dots & \dots & \dots & \dots & \dots & \dots & \dots & \dots \\
  \hline
  7 & CG & CG & \ding{52} & establish & establish & \ding{52} & \ref{n14-7-1} & \ref{s14-7-1} & \ding{52} \\ 
  \hline
  \dots & \dots & \dots & \dots & \dots & \dots & \dots & \dots & \dots & \dots \\
  \hline
  \end{tabular}
  \caption{Article 14 POS table}
  \label{tab:tab-art-14}
  \end{adjustwidth}
\end{table}

\subsection{Mapping of grammatical tags onto modular ontological patterns} \label{sec:app-adapt}
Following coherently what is output from the second step, we are now able to instantiate a concrete representation of the modules by leveraging the pattern depicted in the first step. What we created, in fact, is a \emph{Data Dictionary}, where the essential entities for the ontological representation are collected.
Taking the first row of Table \ref{tab:tab-art-8}, we get the following associations: \emph{Member State} (even if the subject is plural, we always take the singular, as in ontological terms it represents the single entity) as EntityX; \emph{Article8} as ArticleY; \emph{Article-8-MemberState-Compliant} as Article-Y-EntityX-Compliant.

The conceptualisation ends by defining the measure, composed by the verb \emph{designate} and the object referred in \ref{n8-5-1} or \ref{s8-5-1}, depending on the correctness of the NLP response.
The concrete meaning is the following: in order that a \emph{Member State} to be compliant with \emph{Article8}, it must \emph{designate} \ref{n8-5-1}. Obviously, the meaning is the same as in the Directive but we built it starting from the POS that was extracted, which are now mapped on the modular ontological patterns.

\subsection{Implementation of an automated ontology builder}
By executing the code illustrated above in the definition of the method, we get, in general for Article 8, the following extract of the target ontology, presented in Turtle format.

\begin{lstlisting}[caption=Turtle ontological serialisation of the measures of the Directive, breaklines=true]
<http://nas.onto/MemberState> a owl:Class ;
    owl:equivalentClass <http://nas.onto/Article-8-Compliant> 
    
<http://nas.onto/Article-8-Compliant> a owl:Class ;
  owl:equivalentClass [ a owl:Class ;
    owl:intersectionOf ( 
      [ a owl:Restriction ;
        owl:onProperty <http://nas.onto/ensure> ;
        owl:someValuesFrom <http://nas.onto/CA-SinglePOC-HaveAdequateResourcesToCarryOutTasksAssigned > ] 
      [ a owl:Restriction ;
        owl:onProperty <http://nas.onto/pred2> ;
        owl:someValuesFrom <http://nas.onto/Object2> ] ) ] .
      ... 
      [ a owl:Restriction ;
        owl:onProperty <http://nas.onto/predN> ;
        owl:someValuesFrom <http://nas.onto/ObjectN> ] ) ] .
      
\end{lstlisting}

The depicted ontology extract reflects the considerations made within the description of the code. 

\section{Overall evaluation of the method} \label{sec:eval}
Building an ontology for NIS 2 fully relying  on automation was far from simple. Several challenges were encountered.
\paragraph{Full and correct interpretation of sentences}
Strictly following the extraction of sentences via NLP calls and associating the related part of speech to subject, predicate, and object ontological pattern, some information could be lost.
The biggest portion of a sentence is transformed into an object with a low possibility of interpreting its content. 
For example, see the following purposely made-up sentence: \emph{Each Member State shall notify its incidents within 3 months}. 
We correctly get that \emph{its incidents within 3 months} is the object of the sentence but we have no possibility to further extract additional information, for example, 3 months as an ontological data Property, indeed limiting the power ontological representation. 
Even if it were possible, it would be too complex to define a general strategy that includes all possible similar cases.

\paragraph{Full and coherent representation of parts of speech} 
As a design decision, we assumed to rebuild (to ontological language) the articles of the Directive through different and separated modules, each one related to one entity. 
One problem that may arise if automating is the difficulty to associate EntityX with the correct associated measures. Recalling the code \ref{code-onto}, such a problem arises when composing measures from different articles, a task that requires modifying the pre-existent collection of restrictions. 

A similar issue regards the objects since it is difficult to establish a criterion for the name of the entity representing the object itself since it may be composed of many words.

Another problem may arise when the predicate of a sentence is a passive one. In such cases, the object is almost always undetected, and the transformation to the active form is needed. Attempts at transformations are still complex due to the limitations of NLP.

\paragraph{Hierarchical issues}
The automation surely leads to a ``flat hierarchy'' problem.
The extraction of entities and relations lacks critical observation that can be useful for defining a hierarchy among them. In two different contexts that could happen: a) for simple entities; b) for the complex entities that resume with a single word the meaning of an entire object. 

\paragraph{Writing of ontological axioms}
The only axioms we can define are the ones representing each measure we are translating. This happens on a case-by-case basis while parsing the document we are facing. 
It is not possible to add more axioms since we don't know how any of each identified part of speech behaves in the document, not allowing us to make more specific considerations.

\section{Conclusions}\label{sec:conclusions}
The combination of ontologies and NLP aimed at the representation and automation of the reading of security directives turned out to be challenging but, arguably, successful.

The overall task became more manageable by following our method of ontological construction, which required our effort in the first place. Such a method is, in fact, a general contribution of the present paper. It took inspiration from Methontology but rearranged and refocused its steps. In particular, while we deemed the Integration step unnecessary for our purposes because there is no existing vocabulary that can be linked with NIS 2 entities, we expect to complete our method with appropriate instantiations of the Evaluation and Documentation steps in the future, after fully covering the ontology.

If we consider the present work as the first stage of the application of the proposed method, it is clear that some limitations are there. 
Our experiments confirm that modern, forefront NLP techniques struggle with the complex juridical language and nested structuring of the directive. 
The main reason appears to be the number of clauses that are used per sentence and depend on each other, as well as the length of each clause.
Future work will focus on perfecting automation, completing the target ontology, and its companion documentation. 

\section*{Acknowledgments}
Gianpietro Castiglione acknowledges a studentship by Intrapresa S.r.l. and Italian ``Ministero dell’Università e della Ricerca'' (D.M. n. 352/2022).

\bibliography{sample-ceur}

\begin{thebibliography}{16}
\expandafter\ifx\csname natexlab\endcsname\relax\def\natexlab#1{#1}\fi
\providecommand{\url}[1]{\texttt{#1}}
\providecommand{\href}[2]{#2}
\providecommand{\path}[1]{#1}
\providecommand{\DOIprefix}{doi:}
\providecommand{\ArXivprefix}{arXiv:}
\providecommand{\URLprefix}{URL: }
\providecommand{\Pubmedprefix}{pmid:}
\providecommand{\doi}[1]{\href{http://dx.doi.org/#1}{\path{#1}}}
\providecommand{\Pubmed}[1]{\href{pmid:#1}{\path{#1}}}
\providecommand{\bibinfo}[2]{#2}
\ifx\xfnm\relax \def\xfnm[#1]{\unskip,\space#1}\fi
\bibitem[{Gruber(1995)}]{GRUBER1995907}
\bibinfo{author}{T.~R. Gruber},
\newblock \bibinfo{title}{Toward principles for the design of ontologies used
  for knowledge sharing?},
\newblock \bibinfo{journal}{International Journal of Human-Computer Studies}
  \bibinfo{volume}{43} (\bibinfo{year}{1995}) \bibinfo{pages}{907--928}.
  \URLprefix
  \url{https://www.sciencedirect.com/science/article/pii/S1071581985710816}.
  \DOIprefix\doi{https://doi.org/10.1006/ijhc.1995.1081}.
\bibitem[{nis(2023)}]{nisPDF}
\bibinfo{title}{Nis 2 directive}, \bibinfo{year}{2023}. \URLprefix
  \url{https://eur-lex.europa.eu/eli/dir/2022/2555/oj}.
\bibitem[{Fern{\'a}ndez-L{\'o}pez et~al.(1997)Fern{\'a}ndez-L{\'o}pez,
  G{\'o}mez-P{\'e}rez, and Juristo}]{Methontology}
\bibinfo{author}{M.~Fern{\'a}ndez-L{\'o}pez},
  \bibinfo{author}{A.~G{\'o}mez-P{\'e}rez}, \bibinfo{author}{N.~Juristo},
\newblock \bibinfo{title}{Methontology: From ontological art towards
  ontological engineering},
\newblock in: \bibinfo{booktitle}{Proceedings of the Ontological Engineering
  AAAI-97 Spring Symposium Series}, \bibinfo{publisher}{American Asociation for
  Artificial Intelligence}, \bibinfo{year}{1997}. \URLprefix
  \url{https://oa.upm.es/5484/}, \bibinfo{note}{ontology Engineering Group -
  OEG}.
\bibitem[{rdf(2023)}]{rdflib}
\bibinfo{title}{rdflib}, \bibinfo{year}{2023}. \URLprefix
  \url{https://rdflib.readthedocs.io/en/stable/}.
\bibitem[{cla(2023)}]{claucy}
\bibinfo{title}{spacy-clausie library}, \bibinfo{year}{2023}. \URLprefix
  \url{https://spacy.io/universe/project/spacy-clausie}.
\bibitem[{spa(2023)}]{spacy}
\bibinfo{title}{Spacy library}, \bibinfo{year}{2023}. \URLprefix
  \url{https://spacy.io/}.
\bibitem[{git(2023)}]{gitNIS}
\bibinfo{title}{Github resources}, \bibinfo{year}{2023}. \URLprefix
  \url{https://github.com/gianpietroc/nis-ontology}.
\bibitem[{Sawasaki et~al.(2022)Sawasaki, Satoh, and
  Troussel}]{DBLP:conf/semweb/SawasakiST22}
\bibinfo{author}{T.~Sawasaki}, \bibinfo{author}{K.~Satoh},
  \bibinfo{author}{A.~C. Troussel},
\newblock \bibinfo{title}{A use case on {GDPR} of modular-proleg for private
  international law},
\newblock in: \bibinfo{editor}{M.~Navas{-}Loro},
  \bibinfo{editor}{C.~Badenes{-}Olmedo}, \bibinfo{editor}{M.~Koubarakis},
  \bibinfo{editor}{J.~L.~R. Garc{\'{\i}}a}, \bibinfo{editor}{S.~Kirrane},
  \bibinfo{editor}{N.~Mihindukulasooriya}, \bibinfo{editor}{K.~Satoh},
  \bibinfo{editor}{M.~Acosta} (Eds.), \bibinfo{booktitle}{Joint Proceedings of
  the 3th International Workshop on Artificial Intelligence Technologies for
  Legal Documents {(AI4LEGAL} 2022) and the 1st International Workshop on
  Knowledge Graph Summarization (KGSum 2022) co-located with the 21st
  International Semantic Web Conference {(ISWC} 2022), Virtual Event, Hangzhou,
  China, October 23-24, 2022}, volume \bibinfo{volume}{3257} of
  \textit{\bibinfo{series}{{CEUR} Workshop Proceedings}},
  \bibinfo{publisher}{CEUR-WS.org}, \bibinfo{year}{2022}, pp.
  \bibinfo{pages}{1--11}. \URLprefix
  \url{https://ceur-ws.org/Vol-3257/paper1.pdf}.
\bibitem[{Pandit et~al.(2018)Pandit, Fatema, O’Sullivan, and
  Lewis}]{Pandit2018GDPRtEXTG}
\bibinfo{author}{H.~J. Pandit}, \bibinfo{author}{K.~Fatema},
  \bibinfo{author}{D.~O’Sullivan}, \bibinfo{author}{D.~Lewis},
\newblock \bibinfo{title}{Gdprtext - gdpr as a linked data resource},
\newblock in: \bibinfo{booktitle}{Extended Semantic Web Conference},
  \bibinfo{year}{2018}.
\bibitem[{cyb(2023)}]{cyberstrong}
\bibinfo{title}{Cybserstrong platform}, \bibinfo{year}{2023}. \URLprefix
  \url{https://www.cybersaint.io/cyberstrong}.
\bibitem[{Morneau et~al.(2006)Morneau, Mineau, and Corbett}]{SeseiOnto}
\bibinfo{author}{M.~Morneau}, \bibinfo{author}{G.~W. Mineau},
  \bibinfo{author}{D.~Corbett},
\newblock \bibinfo{title}{Seseionto: Interfacing nlp and ontology extraction},
\newblock in: \bibinfo{booktitle}{2006 IEEE/WIC/ACM International Conference on
  Web Intelligence (WI 2006 Main Conference Proceedings)(WI'06)},
  \bibinfo{year}{2006}, pp. \bibinfo{pages}{449--455}.
  \DOIprefix\doi{10.1109/WI.2006.158}.
\bibitem[{Estival et~al.(2004)Estival, Nowak, and Zschorn}]{articleEstival}
\bibinfo{author}{D.~Estival}, \bibinfo{author}{C.~Nowak},
  \bibinfo{author}{A.~Zschorn},
\newblock \bibinfo{title}{Towards ontology-based natural language processing},
\newblock \bibinfo{journal}{Proceedings of NLP-XML 2004}
  (\bibinfo{year}{2004}). \DOIprefix\doi{10.3115/1621066.1621075}.
\bibitem[{Zhang and El-Gohary(2016)}]{doi:10.1061/(ASCE)CP.1943-5487.0000346}
\bibinfo{author}{J.~Zhang}, \bibinfo{author}{N.~M. El-Gohary},
\newblock \bibinfo{title}{Semantic nlp-based information extraction from
  construction regulatory documents for automated compliance checking},
\newblock \bibinfo{journal}{Journal of Computing in Civil Engineering}
  \bibinfo{volume}{30} (\bibinfo{year}{2016}) \bibinfo{pages}{04015014}.
  \DOIprefix\doi{10.1061/(ASCE)CP.1943-5487.0000346}.
\bibitem[{tex(2023)}]{textR}
\bibinfo{title}{Text razor}, \bibinfo{year}{2023}. \URLprefix
  \url{https://www.textrazor.com/}.
\bibitem[{Manning et~al.(2014)Manning, Surdeanu, Bauer, Finkel, Bethard, and
  McClosky}]{manning-EtAl:2014:P14-5}
\bibinfo{author}{C.~D. Manning}, \bibinfo{author}{M.~Surdeanu},
  \bibinfo{author}{J.~Bauer}, \bibinfo{author}{J.~Finkel},
  \bibinfo{author}{S.~J. Bethard}, \bibinfo{author}{D.~McClosky},
\newblock \bibinfo{title}{The {Stanford} {CoreNLP} natural language processing
  toolkit},
\newblock in: \bibinfo{booktitle}{Association for Computational Linguistics
  (ACL) System Demonstrations}, \bibinfo{year}{2014}, pp.
  \bibinfo{pages}{55--60}. \URLprefix
  \url{http://www.aclweb.org/anthology/P/P14/P14-5010}.
\bibitem[{Musen(2015)}]{DBLP:journals/aimatters/Musen15}
\bibinfo{author}{M.~A. Musen},
\newblock \bibinfo{title}{The prot{\'{e}}g{\'{e}} project: a look back and a
  look forward},
\newblock \bibinfo{journal}{{AI} Matters} \bibinfo{volume}{1}
  (\bibinfo{year}{2015}) \bibinfo{pages}{4--12}. \URLprefix
  \url{https://doi.org/10.1145/2757001.2757003}.
  \DOIprefix\doi{10.1145/2757001.2757003}.

\end{thebibliography}

\appendix
\section*{Appendix}

\section*{Article objects \& Acronym tables}
\def\arraystretch{0.6}
\begin{longtable}{|p{7cm}|p{6cm}|}
\hline
\multicolumn{2}{|c|}{\textbf{Article:} 8 } \\
\hline
\multicolumn{2}{|c|}{\textbf{\textbf{Item 8.1}}} \\
\hline
\begin{enumerate}[label=\textbf{N8.1}]
    \item one or more competent authorities responsible for cybersecurity and for the supervisory tasks referred to in Chapter VII (competent authorities) \label{n8-1-1}
  \end{enumerate} & 
  \begin{enumerate}[label=\textbf{S8.1}]
    \item one or more competent authorities responsible for cybersecurity \label{s8-1-1}
  \end{enumerate} \\
\hline
\multicolumn{2}{|c|}{\textbf{\textbf{Item 8.2}}} \\
\hline
 \begin{enumerate}[label=\textbf{N8.2}]
    \item the implementation of this Directive at national level\label{n8-2-1}
  \end{enumerate} &
  \begin{enumerate}[label=\textbf{S8.2}]
    \item the implementation of this Directive at national level \label{s8-2-1}
  \end{enumerate} \\
\hline
\multicolumn{2}{|c|}{\textbf{\textbf{Item 8.3}}} \\
\hline
  \begin{enumerate}[label=\textbf{N8.3.{\arabic*}}, , leftmargin=1.2cm]
    \item a  single  point  of  contact \label{n8-3-1}
    \item that competent authority shall also be the single point of contact for that Member State \label{n8-3-2} 
  \end{enumerate} & 
  \begin{enumerate}[label=\textbf{S8.3.{\arabic*}}, leftmargin=1.1cm]
    \item a single point of contact \label{s8-3-1}
    \item only one competent authority \label{s8-3-2}
  \end{enumerate} \\
\hline
\multicolumn{2}{|c|}{\dots} \\
\hline
\multicolumn{2}{|c|}{\textbf{\textbf{Item 8.5}}} \\
\hline
\begin{enumerate}[label=\textbf{N8.5} , leftmargin=1.2cm]
        \item that  their  competent  authorities  and  single  points  of  contact  have  adequate  resources  to carry  out,  in  an  effective  and  efficient  manner \label{n8-5-1}
    \end{enumerate} &
    \begin{enumerate}[label=\textbf{S8.5}]
        \item a) that  their  competent  authorities  and  single  points  of  contact  have  adequate  resources  to carry  out,  in  an  effective  and  efficient  manner | b)adequate resources \label{s8-5-1}
    \end{enumerate} \\ 
\hline
\multicolumn{2}{|c|}{\textbf{\textbf{Item 8.6}}}  \\
\hline
\begin{enumerate}[label=\textbf{N8.6.{\arabic*}}, leftmargin=1.2cm]
        \item  the  Commission  without  undue  delay  of  the  identity  of  the  competent  authority referred  to  in  paragraph  1  and  of  the  single  point  of  contact  referred  to  in  paragraph  3, of  the  tasks  of  those  authorities, and  of  any  subsequent  changes  thereto \label{n8-6-1}
    \end{enumerate} &
    \begin{enumerate}[label=\textbf{S8.6.{\arabic*}}, leftmargin=1.2cm]
        \item the commission \label{s8-6-1}
    \end{enumerate} \\
\hline
\multicolumn{2}{|c|}{\dots} \\
\hline
\caption{Article 8 objects tabulation}
\end{longtable}

\begin{longtable}{|p{7cm}|p{6cm}|}
\hline
\multicolumn{2}{|c|}{\textbf{Article:} 14 } \\
\hline
\multicolumn{2}{|c|}{\textbf{\textbf{Item 14.1}}} \\
\hline
  \begin{enumerate}[label=\textbf{N14.1.\arabic*}, leftmargin=1.5cm]
    \item PASSIVE - NONE \label{n14-1-1}
  \end{enumerate} &
  \begin{enumerate}[label=\textbf{S14.1.\arabic*}, leftmargin=1.3cm]
    \item PASSIVE - NONE \label{s14-1-1}
  \end{enumerate} \\
\hline
\multicolumn{2}{|c|}{\textbf{\textbf{Item 14.2}}} \\
\hline
\begin{enumerate}[label=\textbf{N14.2.\arabic*}, leftmargin=1.4cm]
    \item its tasks on the basis of biennial work programmes referred to in paragraph 7 \label{n14-2-1}
  \end{enumerate} &
  \begin{enumerate}[label=\textbf{S14.2.\arabic*}, leftmargin=1.3cm]
    \item its tasks on the basis of biennial work programmes referred to in paragraph 7 \label{s14-2-1}
  \end{enumerate} \\
\hline
\multicolumn{2}{|c|}{\textbf{\textbf{Item 14.3}}} \\
\hline
\begin{enumerate}[label=\textbf{N14.3.\arabic*}, leftmargin=1.4cm]
    \item of representatives of Member States, the Commission and ENISA \label{n14-3-1}
    \item in the activities of the Cooperation Group as an observer \label{n14-3-2} 
    \item in the activities of the Cooperation Group in accordance with Article 47(1) of that Regulation. \label{n14-3-3}
  \end{enumerate} &
  \begin{enumerate}[label=\textbf{S14.3.\arabic*}, leftmargin=1.4cm]
    \item NONE \label{s14-3-1}
    \item NONE \label{s14-3-2}
    \item NONE \label{s14-3-3}
  \end{enumerate} \\
\hline
\multicolumn{2}{|c|}{\dots} \\
\hline
\multicolumn{2}{|c|}{\textbf{\textbf{Item 14.7}}} \\
\hline
\begin{enumerate}[label=\textbf{N14.7.\arabic*}, leftmargin=1.4cm]
    \item a work programme in respect of actions to be undertaken to implement its objectives and tasks \label{n14-7-1}
  \end{enumerate} &
  \begin{enumerate}[label=\textbf{S14.7.\arabic*}, leftmargin=1.4cm]
    \item a work programme in respect of actions to be undertaken to implement its objectives and tasks \label{s14-7-1}
  \end{enumerate} \\
\hline
\multicolumn{2}{|c|}{\dots} \\
\hline
\caption{Article 14 objects tabulation}
\end{longtable}

\end{document}